\documentclass[11pt]{article}
\usepackage[utf8]{inputenc}
\usepackage[T1]{fontenc}
\usepackage{lmodern}
\usepackage[margin=1in]{geometry}
\usepackage{amsmath,amssymb}
\usepackage{graphicx}
\usepackage{booktabs}
\usepackage{tabularx}
\usepackage{multirow}
\usepackage{array}
\usepackage{caption}
\usepackage{newunicodechar}
\usepackage[hidelinks]{hyperref}
\newunicodechar{κ}{\ensuremath{\kappa}}
\newunicodechar{≥}{\ensuremath{\geq}}
\newunicodechar{≤}{\ensuremath{\leq}}
\newunicodechar{×}{\ensuremath{\times}}
\newunicodechar{±}{\ensuremath{\pm}}
\newunicodechar{∩}{\ensuremath{\cap}}
\newunicodechar{∪}{\ensuremath{\cup}}
\newunicodechar{−}{\ensuremath{-}}
\newunicodechar{·}{\ensuremath{\cdot}}
\newunicodechar{→}{\ensuremath{\rightarrow}}
\newunicodechar{≈}{\ensuremath{\approx}}
\newunicodechar{†}{\textdagger}
\newunicodechar{‐}{-}
\newunicodechar{‑}{-}
\newunicodechar{–}{\textendash}
\newunicodechar{—}{\textemdash}
\newunicodechar{’}{'}
\newunicodechar{‘}{'}
\newunicodechar{“}{``}
\newunicodechar{”}{''}
\newunicodechar{…}{\ldots}
\begin{document}

\begin{center}
{\LARGE\bfseries Simultaneous hyperkinetic movement disorders phenotyping: a cross-cohort pediatric transfer study using routine videos, markerless pose estimation and a tabular foundation model}\\[1.2em]
{\large Laura Cif \textsuperscript{1,~2,}, Diane Demailly \textsuperscript{2,3*}, Zohra Souei \textsuperscript{2,4}*, Muhammad Mushhood Ur Rehman \textsuperscript{5}, Juan Dario Ortigoza Escobar \textsuperscript{6,7,8}, Mayté Castro Jiménez \textsuperscript{1} , Cécile A. Hubsch\textsuperscript{1}, Sophie Huby \textsuperscript{9}, Morgan Dornadic \textsuperscript{9}, Gun-Marie Hariz \textsuperscript{10}, Eduardo M. Moraud\textsuperscript{11}, Jocelyne Bloch \textsuperscript{11,12}, Gabriella A. Horvàth \textsuperscript{13} and Xavier Vasques \textsuperscript{2}}
\end{center}

{\footnotesize\begin{flushleft}
\textsuperscript{1} Service of Neurology, Department of Clinical Neurosciences, Lausanne University Hospital (CHUV) and University of Lausanne (UNIL), Lausanne, Switzerland\\
\textsuperscript{2} Institut du Neurone, Montferrier sur Lez, France\\
\textsuperscript{3} Department of Neurology, Clinique Beau Soleil, Institut Mutualiste Montpelliérain, Montpellier, France\\
\textsuperscript{4} Department of Neurosurgery, Military University Hospital of Sfax, Tunisia.\\
\textsuperscript{5} University of Edinburgh, Scotland\\
\textsuperscript{6} Movement Disorders Unit, Pediatric Neurology Department, Institut de Recerca, Hospital Sant Joan de Déu, Barcelona, Spain\\
\textsuperscript{7} European Reference Network for Rare Neurological Diseases (ERN-RND), Barcelona, Spain\\
\textsuperscript{8} U-703 Centre for Biomedical Research on Rare Diseases (CIBER-ER), Instituto de Salud Carlos III, Barcelona, Spain\\
\textsuperscript{9} Department of Neurology, CHU Montpellier, 34295, Montpellier, France\\
\textsuperscript{10} Department of Clinical Neuroscience, Umeå University, Umeå, Sweden\\
\textsuperscript{11} Defitech Center for Interventional Neurotherapies (NeuroRestore), University Hospital Lausanne and Ecole Polytechnique Fédérale de Lausanne, Lausanne, Switzerland\\
\textsuperscript{12} Department of Neurosurgery, Lausanne University Hospital (CHUV) and University of Lausanne (UNIL), Lausanne, Switzerland.\\
\textsuperscript{13} Department of Pediatrics, British Columbia Children's Hospital, Vancouver, British Columbia, Canada\\[0.6em]
\noindent$^{*}$Correspondence: Laura Cif, \href{mailto:lauracif@institutduneurone.fr}{lauracif@institutduneurone.fr}
\end{flushleft}}


\begin{abstract}
\textbf{Background:} Accurate recognition of movement disorders (MDs) phenomenology remains a demanding tasks in clinical neurology. In pediatric practice, this challenge is compounded by mixed and evolving motor presentations, inter-examiner expertise variability, and the scarcity of clinical and of structured digital tools validated across age groups. A child with combined MDs may receive different phenomenological diagnostics from different examiners, with direct consequences for etiological guidance, treatment planning, neuromodulation candidacy, and longitudinal monitoring. Video-based examination is already central to expert practice, but its systematic analysis and use has remained, due to frequently limited human resources, out of reach for many teams.

\textbf{Objective:} To develop and externally test a video-based framework for simultaneous detection of hyperkinetic MDs phenomenologies: dystonia, tremor, myoclonus, chorea, athetosis, ballismus, stereotypies, and tics using routine clinical recordings, with explicit testing of external, cross-cohort transfer from adult to pediatric populations.

\textbf{Methods:} In this proof-of-concept study, the framework combines markerless pose estimation, kinematic descriptors, and a pretrained fondation model. A shared predictive backbone was developed on 21 adults with confirmed hyperkinetic MDs and 4 healthy controls assessed under a standardized protocol. External validation was performed on an independent external cohort: a real-world pediatric sample (n=12, monogenic combined MDs). For the external dataset, the backbone was deployed without retraining; lightweight calibration adjusted only the final subject-level decision step using a small labeled subset of patients selected by clinicians as representative of the cohort\textquotesingle s phenotypic range.

\textbf{Results:} After local calibration of the decision layer on the clinician-selected subset, performance improved consistently on the held-out pediatric patients (n=7): Hamming accuracy rose from 0.804 to 0.839 and the Jaccard index from 0.548 to 0.633. This calibrated performance was preserved, and the Jaccard index further improved, when the evaluation was restricted to the phenomenologies with more definite clinician agreement (Hamming accuracy 0.9, Jaccard index 0.786), indicating that the gains did not rest on the least-reliable labels.

\textbf{Conclusion:~}A two-stage strategy, a reusable predictive backbone combined with lightweight, site-specific adaptation, may offer a practical and scalable route to routine-video-based digital phenotyping of MDs across age groups. A shared backbone trained on standardized adult recordings transferred clinically useful phenotyping signal to an independent, real-world pediatric cohort without retraining. Local calibration of the decision layer on a small, clinician-selected subset substantially improved performance, most clearly for well-represented phenomenologies, while clinical-grade reliability across the full phenotypic spectrum remains to be confirmed in larger and more heterogeneous cohorts.
\end{abstract}

\section{Introduction}

Movement disorders (MDs) encompass a heterogenous group of neurological conditions including hypo- and hyperkinetic MDs, both in adults and pediatric populations, with age and etiology-related specificities. Defined by the presence of abnormal involuntary movements, hyperkinetic MDs (HMDs) including dystonia, tremor, myoclonus, chorea, athetosis, ballismus, stereotypies and tics frequently co-occur within the same patient, fluctuate and evolve over time. In everyday practice, video-based examination remains central to the recognition of these phenomenologies, yet this process is inherently demanding: it requires substantial specialist expertise, takes time, and remains affected by subjectivity \textsuperscript{1--4}. The clinical stakes of misclassification are high: inappropriate phenotyping can lead to delayed diagnosis, suboptimal pharmacological management and particularly in the context of neuromodulation, incorrect patient and target selection or inappropriate deep brain stimulation (DBS) administration. These challenges are particularly relevant in pediatric neurology: neurodevelopmental, neurometabolic, neurodegenerative disorders or monogenic MD often present with mixed, variable and evolving phenomenologies that do not map cleanly onto adult-derived taxonomies \textsuperscript{2}. Longitudinal tracking of motor phenomenology is needed not only for diagnosis but also for treatment planning across transition from pediatric to adult care and, for outcomes evaluation. Yet there is currently no validated, scalable digital tool for structured multi-symptom assessement of that phenomenologies across age groups and recording conditions.

To address this clinical complexity, we previously developed and preliminarily validated the Combined Dystonia-Scale for Assessment of the Motor Phenotype (CODY-SAMP), a comprehensive clinical scale for the assessment of combined and mixed MDs across pediatric and adult populations. CODY-SAMP provides a standardized framework, indexing and rating the presence, intensity, and distribution of each HMD phenomenology and across different conditions, enabling reproducible serial assessement. In parallel with this clinical effort, we developed a video-based deep learning pipeline for binary (symptom present or absent) and multi-symptom recognition of eight HMD phenomenologies that may represent clinically meaningful therapeutic targets, including in the context of neuromodulation \textsuperscript{5}. Initial supervised classification models were trained in a single-center adult cohort and yielded encouraging results. However, promising performance within a development cohort does not establish neither generalizability nor clinical usefulness. Before such an approach can become relevant in practice, it must show that it can generalize beyond the original site and beyond the original patient population, including transfer from adults to pediatric and heterogeneous real-world clinical cohorts.

Recent systematic reviews have documented rapid growth in video-based analysis with artificial intelligence (AI) for MDs, yet have also identified persistent critical limitations: most published systems target a single symptom (tremor or tics), operate under controlled acquisition conditions, or are validated only within their development cohort \textsuperscript{6,7}. Real clinical environments are more demanding, videos are acquired with heterogenous cameras, under variable framing and lighting, and in patients who may not cooperate with standardized protocols. A clinically useful system therefore needs to cope with variability in age, phenotype composition, recording conditions and annotation structure across cohorts.

Markerless pose estimation is especially attractive because it can transform standard clinical videos into structured movement trajectories without specialized motion-capture hardware, thereby preserving compatibility with routine workflows \textsuperscript{1,6,7}. Classical machine-learning methods (such as logistic regression, random forests, support vector machines and gradient-boosted trees) remain powerful in this setting, but require repeated model selection and dataset-specific tuning which can be a significant practical burden when cohorts are small, labels are imbalanced, and recording conditions differ across sites \textsuperscript{8--10}. These limitations are particularly important in rare or heterogeneous clinical populations, where annotated datasets are expensive to produce and difficult to replicate across institutions.~Tabular foundation models offer a different strategy. Rather than learning each task from scratch, they are pretrained on very large collections of tabular problems so as to acquire a transferable inductive bias for structured prediction \textsuperscript{8--10}. Applied to clinical video phenotyping, a pretrained tabular foundation model may provide a reusable predictive backbone less dependent on full retraining at each new site which is precisely the bottleneck that limits real-word deployment of video-AI tools in these specific populations. From a clinical perspective, the practical goal is not simply to maximize performance within a single dataset, but to reduce the burden of re-engineering a complete model every time local conditions change. What matters is whether a model trained in one setting can still provide useful signal in another, and whether the remaining mismatch can be addressed through lighter local adaptation of the decision layer rather than full redevelopment.

We therefore evaluated a video-based framework for subject-level simultaneous multi-symptom detection from clinical recordings, explicitly designed to test transfer across external cohorts. The system combines markerless pose estimation, kinematic feature extraction and a pretrained tabular foundation model, and targets the aforementioned HMD phenomenologies. Importantly, the framework separates a shared predictive backbone, developed from standardized adult video recordings, from a dataset specific, adaptive subject-level decision layer that can be adjusted locally in pediatric or adult cohorts without retraining the core model. We evaluated this strategy across one real-world independent external pediatric dataset. Our aim was twofold: first, to determine whether routine clinical videos contain sufficiently robust structured information to support clinically meaningful multi-symptom phenotyping across a pediatric cohort; and second, to assess whether a shared backbone trained in adults can serve as a practical basis for transfer to pediatric populations.

\section{Methods}

\subsection{Study design, participants, video acquisition and ethics}

This proof-of-concept study evaluated a framework for subject-level multi-symptom phenotyping of HMDs from clinical video recordings (\textbf{Figure 1}). The overall workflow comprised three stages: (i) model training on a first cohort with annotated standardized adult videos, (ii) external inference on one independent sample without backbone retraining, and (iii) dataset-specific calibration of the final subject-level decision step. The target output at each stage was the presence or absence, at subject level, of the eight HMDs phenomenologies indexed by CODY-SAMP, assessable on video recordings, although CODY-SAMP includes a broader clinical assessment of motor features, such as parkinsonism, ataxia or dysarthria, not modeled in the present study.

The training cohort was recruited at Beau Soleil Clinic, Montpellier, France, and included twenty-five participants, twenty-one patients and four healthy controls. Briefly, all participants underwent video-recorded examination and were clinically assessed using CODY-SAMP (\textbf{Supplementary Material S1}). Each video recording was rated independently by two clinicians, yielding 50 full-examination labelling. In addition, the same 25 participants contributed condition-specific videos corresponding to rest, posture and action tasks specifically involving the upper limbs, yielding 75 additional video-derived files labelled by one clinician. Altogether, 125 labeled video-derived files were used for annotation, feature extraction and downstream model training. For the condition-specific recordings, rest and posture lasted 10 seconds each, whereas action recordings lasted 20 seconds. The CODY-SAMP-related full video sessions lasted approximately 12.5 minutes when all tasks were completed and cooperation was adequate.

Evaluation was then performed on one independent external dataset having as common denominator referral and follow-up for combined MDs including dystonia. The dataset consisted of 12 pediatric patients recorded during routine clinical practice outside the standardized protocol and therefore representing a more difficult real-world setting, with greater variability in framing, camera positioning and overall video quality. Ethics approval was obtained from the University Hospital Montpellier, France (IRB-MTP\_2020\_09\_202000565), and Beau Soleil Clinic, Montpellier, France (CESSRESS 22075132 Bis and CNIL n°2238428). Informed consent for study participation and video recording was obtained from all patients or their legal guardians and the procedures complied with the Declaration of Helsinki, 1975, as revised in 2000.~For these external datasets the developed framework applied site calibration of the final decision step (when specified) and generated subject level HMD multi-symptom predictions.

\begin{figure}[htbp]
\centering
\includegraphics[width=\textwidth,keepaspectratio]{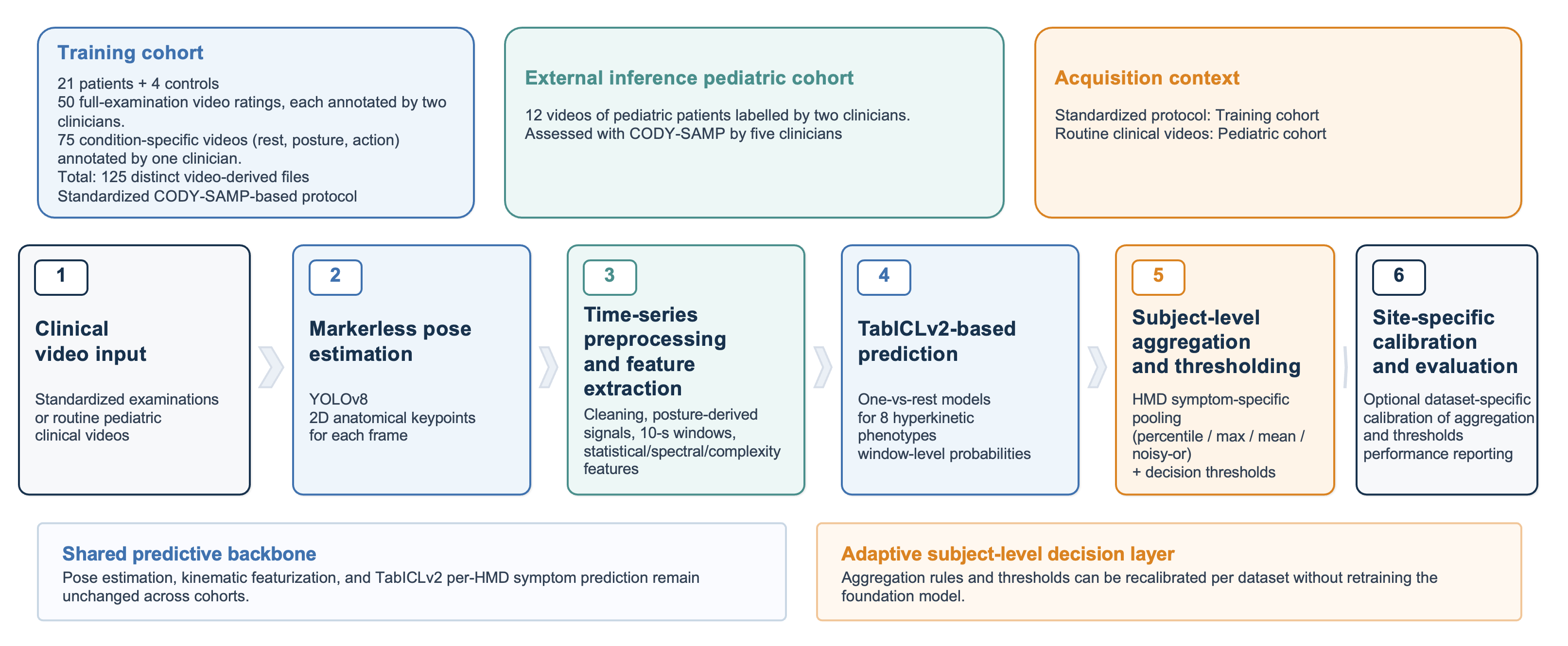}
\caption*{\textbf{Figure 1.} Study cohorts and video-based phenotyping pipeline for HMDs. Top: the training cohort (21 patients with combined HMDs and 4 controls, standardized CODY-SAMP protocol) and the external pediatric inference cohort (12 patients with monogenic combined MDs, routine clinical videos), illustrating the contrast in acquisition context between training and inference. Bottom: the six-step pipeline, in which a shared predictive backbone (markerless pose estimation, kinematic featurization and TabICLv2 per-HMD symptom prediction; steps 1--4) is held fixed across cohorts, while an adaptive subject-level decision layer (aggregation rules and decision thresholds; steps 5--6) can be recalibrated per dataset without retraining the foundation model.}
\end{figure}

\subsection{Clinical assessment and video annotation procedure}

\textbf{Training cohort}

All raters were pediatric and adult MD specialists. Before scoring with the CODY-SAMP, they reviewed the scale manual and the standardized video protocol instructions. A calibration session based on example videos was used to harmonize interpretation of the phenomenologies and associated scoring rules. Ratings were performed independently, and raters were blinded to one another's evaluations and to the study hypotheses. Complete blinding to the underlying clinical diagnosis was not always achievable, because some phenomenologies are recognizable on video from motor presentation alone, acknowledged as a potential source of structured bias in the reference phenomenologies. The standardized examination videos were independently annotated by two raters. Video segments were labeled for each of the eight possible HMD phenomenologies using a predefined coding rule: present (1), absent (0) or uncertain (2). Ten-second temporal windows were then defined from clinician annotations specifying start and end times. This window duration was chosen as a practical compromise: it was long enough to provide stable movement descriptors, but short enough to capture the temporal fluctuations typical of HMDs. Each possible HMD phenomenology was evaluated across three examination contexts: rest, posture and action. Healthy controls underwent neurological examination to exclude clinically relevant movement abnormalities. In the model training pipeline, labels marked as uncertain were treated as unknown for the corresponding phenomenology and were excluded from model fitting and threshold tuning for that label. Only explicitly positive or explicitly negative labels contributed to training. This strategy allowed clinically ambiguous segments to remain documented, while avoiding the need to force them into an artificial present (1) or absent (0) label.

\textbf{External cohort}

For the external cohort, the assessment framework differed according to the dataset. The pediatric dataset was rated with CODY-SAMP by five MD experts. The clinical examination recordings were not annotated 10 second window-wise but phenomenologies were available for each subject and for each assessor from assessment by the CODY-SAMP. Rather than forcing all expert ratings into a single simplified consensus phenomenology label, we retained several clinically meaningful definitions for a HMD symptom label present (i.e.~1) or absent (i.e.~0) based on the combination of the different assessors evaluation. In practice, this means that model performance was examined using different ways of turning expert ratings into HMD phenomenology retained labels: one based on the prevailing expert opinion, and others restricted to those labels with clearer expert agreement. This was done to reflect that disagreement between raters is part of the clinical reality of MD phenomenology rather than simple annotation noise.

\subsection{Pose estimation and kinematic feature extraction}

Full technical details are provided in \textbf{Supplementary Material S2}. Briefly, all videos were converted into frame-wise markerless pose trajectories. For each frame, 2D anatomical keypoints were extracted using a YOLOv8-based pose-estimation pipeline, generating a structured skeletal representation compatible with routine video acquisition. The pose-derived trajectories were then organized as multivariate time series. For inference data, preprocessing included removal of fully empty rows, optional filtering of sparse rows, and reconstruction of missing distance signals from available x-y coordinates when needed. Additional posture-related signals were derived when sufficient trunk landmarks were available \textsuperscript{5}, including descriptors related to shoulder and hip alignment, trunk angle, torsion and postural asymmetry, preprocessing intended to improve robustness to variability in real-world recordings while preserving clinically interpretable biomechanical structure.

Feature extraction was performed on fixed temporal windows. In the training cohort, 10-second windows were aligned to clinician-defined annotated segments. In the external pediatric cohort, a sliding-window procedure was used (300 frames, stride 150 frames at 30 fps, corresponding to 10-second windows with 50\% overlap). Within each window, for each signal a core set of 19 interpretable kinematic features was computed spanning distributional, temporal, spectral, and complexity domains, including measures of central tendency, energy dominant spectral peak frequency and amplitude, Higuchi fractal dimension, and permutation entropy \textsuperscript{11} \textsuperscript{12} with additional posture-derived descriptors when available (\textbf{Supplementray Matrial S2}). These features were selected because they provide reusable quantitative movement summaries that could, in future applications, be related to other clinical, electrophysiology or imaging data.~To reduce sensitivity to features scale differences and outliers, normalization was applied at the signal level before feature computation, using median- and interquartile-range-based scaling. The resulting feature tables formed the structured input to the downstream pretrained tabular foundation model \textsuperscript{13}.

\subsection{Foundation-model-based HMD prediction pipeline}

The pose-derived kinematic feature tables were used as input to a pretrained TabICLv2 tabular foundation model deployed in a supervised clinical prediction pipeline \textsuperscript{10} \textsuperscript{9} \textsuperscript{8}. The reason for using a pretrained tabular foundation model rather than training a conventional classifier from scratch was pragmatic as well as methodological. Once movement has been transformed into structured kinematic descriptors, the prediction task becomes tabular. In that setting, a foundation model offers the possibility of reusing a shared predictive backbone across cohorts, rather than rebuilding a new model from the beginning each time external conditions change \textsuperscript{10} \textsuperscript{9}. Because the clinical target was multi-symptom rather than mutually exclusive, prediction was formulated as eight separate binary classification tasks, one for each symptom. This design was chosen because it allows each phenomenology to have its own decision boundary, prevalence profile and later threshold adjustment. For each symptom, the model produced a probability presence for each analyzed time window.

To limit the effect of class imbalance, absent examples were subsampled relative to present examples within each symptom specific training set, while preserving all available control absents. During development, out-of-fold window-level probabilities were generated using subject-grouped cross-validation, so that all windows from the same individual remained in the same fold and information leakage was avoided. These out-of-fold probabilities were then used for downstream subject-level aggregation and threshold tuning.

\subsection{Subject-level phenomenology aggregation and source-specific calibration}

The model produced one probability per window and per phenomenology. These repeated window-level probabilities were then converted into one subject-level score per phenomenology by combining them across all windows belonging to the same subject. Because different phenomenologies may have different temporal expression patterns, this combination step was not assumed to be identical across symptoms. Instead, the aggregation rule for each symptom was selected during development from a predefined set of candidate rules including percentile-based pooling, maximum pooling, mean pooling and noisy-or aggregation.~Concretely, the aggregation step is what turns a sequence of approximatively 60 window-level probabilities for a given patient into one subject-level score per phenomenology. Different aggregation rules suit different clinical patterns: a mean rule emphasizes symptoms expressed continuously across the recording (e.g., tremor), whereas a maximum or high-percentile rule is better suited to phenomenologies expressed as brief intermittent bursts (e.g., myoclonus). The optimal rule per phenomenology was therefore selected from the data rather than imposed \emph{a priori.}

The next step was to convert these subject-level scores into present /absent subject-level predictions. This required selecting a threshold for each label (present-1, absent-0). Threshold selection was also performed separately for each phenomenology using subject-grouped cross-validation on the training cohort. The optimization criterion combined subject-level predictive performance with penalties intended to discourage excessive positive calling and to preserve conservative behavior in healthy controls.

For the external dataset, the aggregation rules and thresholds derived from the training cohort served as the default deployment configuration. To evaluate adaptation to a new clinical setting, an additional local calibration step could then be applied: the final decision step was adjusted locally, using a small labeled subset of the target dataset, by refining how repeated window-level probabilities were combined at the subject level and where the positivity thresholds were placed. The shared predictive backbone itself was not retrained, meaning no heavy computation and no need for machine-learning expertise within the local clinical team. In an envisaged clinical-deployment context, this calibration step would be performed once when the model is first applied in a new clinical setting, analogous to the commissioning of a new clinical device, and revisited only if the local recording protocol changes substantially. It is carried out jointly with the local clinical team, who provide the ground-truth labels for the small calibration subset. In the pediatric dataset, calibration was formulated as a multi-symptom optimization problem that included penalties for overprediction and optional constraints related to clinically important phenomenologies. In the pediatric cohort, this calibration subset comprised five of the twelve patients, selected by the movement-disorder specialists as collectively representative of the cohort's phenotypic spectrum; the remaining seven patients were held out and used exclusively for external evaluation. Because the calibration and held-out sets were disjoint, performance on the held-out patients provides the primary estimate of external validity, whereas comparison with the calibration subset was used to verify that the local tuning did not overfit.

\subsection{Label definitions and inter-rater analyses}

Ground-truth labels (symptom present 1, vs absent 0) for external evaluation were derived from multi-rater clinician annotations using several clinically meaningful label definitions. The purpose was not to create unnecessary complexity, but to reflect the fact that expert agreement varies across phenomenologies and across datasets. In practical terms, we evaluated the model in two broad ways: first, using the main present/absent clinical label definition for each dataset; and second, using gradually stricter definitions as well as data restricted to labels with higher expert agreement only. In this latest scenario, data with lower expert agreements were discarded. The operational definitions for the labels present or absent used in the study are summarized in \textbf{Table 1}.

For pediatric the dataset assessed by five raters, binary labels definition (present 1, absent 0) was gradual, with ~main present/absent (≥3/5) (meaning that a phenomenology was considered present when at least three of five raters scored it as present, otherwise labelled as absent) main present/absent (≥4/5)~and~main present/absent (5/5) respectively, generated for sensitivity analyses. In addition, we created stricter evaluation settings that retained clear present and clear absent cases but left intermediate expert disagreement unscored rather than forcing them into a present (1) or absent (0) label. In the manuscript, these are referred to as the restrictive agreement-based evaluation. Clinically, they can be understood as analyses restricted to cases with clear expert agreement.

Dataset-level descriptive analyses included phenotype prevalence, vote distributions for the different HMD phenomenology labels (presence or absence) and the proportion of labels affected by disagreement. Inter-rater agreement was quantified separately for each dataset and phenomenology using appropriate categorical agreement statistics.

\begin{table}[htbp]\centering\footnotesize
\caption*{\textbf{Table 1.} Label definitions used for evaluation in the pediatric cohort.}
\renewcommand{\arraystretch}{1.2}
\begin{tabularx}{\textwidth}{@{}p{3.1cm}XXX@{}}
\toprule
\textbf{Label definition} & \textbf{Present rule} & \textbf{Absent rule} & \textbf{Intermediate / ambiguous cases} \\
\midrule
Main present/absent (≥3/5) & Present (1) if at least 3 of 5 raters marked the symptom present & Absent (0) otherwise & None; all cases forced to binary present (1) or absent (0) \\
Main present/absent (≥4/5) & Present (1) if at least 4 of 5 raters marked the symptom present & Absent (0) otherwise & None; all cases forced to binary present (1) or absent (0) \\
Main present/absent (5/5) & Present (1) only if all 5 raters marked the symptom present & Absent (0) otherwise & None; all cases forced to binary present (1) or absent (0) \\
Restrictive agreement-based (≥3/5; no absent) & Present if at least 3 of 5 raters marked the symptom present & Absent if none of the 5 raters marked it present & All remaining cases excluded from evaluation \\
Restrictive agreement-based (≥4/5; no absent) & Present if at least 4 of 5 raters marked the symptom present & Absent if none of the 5 raters marked it present & All remaining cases excluded from evaluation \\
Restrictive agreement-based (5/5; no absent) & Present only if all 5 raters marked the symptom present & Absent if none of the 5 raters marked it present & All remaining cases excluded from evaluation \\
\bottomrule
\end{tabularx}
\end{table} Definitions used to derive subject-level evaluation labels from five-expert annotations in the pediatric cohort. The table summarizes the positivity rule, negativity rule and handling of intermediate cases across the majority-based and restricted agreement-based definitions used in the study. These definitions enabled sensitivity analyses across different levels of expert agreement.

\subsection{Model evaluation protocol and performance metrics}

Model evaluation was conducted at the subject level, the clinically relevant unit of analysis. For each patient, the inference pipeline produced an eight-element present/absent vector, one decision per HMD phenomenology, which was compared against the corresponding eight-element clinician-derived vector under each label definition. Two complementary multi-label metrics were used. To make the metrics concrete, consider a patient for whom clinicians recorded two phenomenologies as present, dystonia and myoclonus, while the model predicted those two plus tremor; the remaining five phenomenologies were absent and were correctly predicted absent. Hamming accuracy counted, per patient, the proportion of the eight individual labels correctly classified. In this example,~seven of the eight labels are concordant~(the five truly absent phenomenologies plus the correctly detected dystonia and myoclonus) and~one is discordant~(tremor is predicted positive but is clinician-negative), giving a Hamming accuracy of 7/8 = 0.875. Jaccard index~restricted the comparison to positive labels only: it is the size of the intersection between the set of model-predicted positive phenomenologies and the set of clinician-defined positive phenomenologies, divided by the size of their union. In the same example, the intersection is \{myoclonus, dystonia\} (size 2) and the union is \{myoclonus, dystonia, tremor\} (size 3), giving Jaccard = 2/3 = 0.667. Jaccard is informative precisely where Hamming can be misleading: in a population where most labels are absent for most patients, true negatives inflate Hamming, whereas Jaccard isolates the agreement on what the clinicians considered clinically present. The two metrics are therefore complementary: Hamming captures partial concordance across all eight labels; Jaccard focuses on the positive phenomenologies, the actionable subset for clinical decision-making. In addition, phenomenology-specific confusion summaries (true positives, true negatives, false positives, false negatives) were generated for each of the eight phenomenologies and reported separately for each label definition (main present/absent definition; majority-based definitions in pediatric dataset; restricted definitions retaining only clear-agreement labels). In analyses using a restricted label definition, labels left unresolved because of expert disagreement were excluded from scoring for the corresponding patient--label pair, and the number of retained labels was recorded. During model training, the per-phenomenology aggregation rule (the function used to combine the \textasciitilde60 window-level probabilities of a given patient into one subject-level score) and the corresponding decision threshold were tuned using patient-grouped cross-validation, so that all windows from the same patient were assigned to a single fold, preventing within-patient information leakage between training and validation. In the external cohorts, baseline (uncalibrated) and locally calibrated deployments were evaluated under the same framework, enabling direct comparison of patient-level performance before and after adjustment of the final decision step. In the present comparison, the uncalibrated baseline applied a generic, site-agnostic decision rule, a 0.5 probability threshold with high-percentile (p95) temporal aggregation, rather than the development-derived configuration, so that the contrast with the calibrated deployment isolates the contribution of local decision-layer calibration. For the pediatric cohort, patient-level performance is reported primarily on the seven held-out patients, which provide the external-validation estimate, and, for completeness, on the full cohort of twelve patients; the five calibration patients are reported separately to document that local tuning did not lead to overfitting.

\subsection{Statistical analysis}

All analyses were conducted at the patient level unless otherwise specified. Dataset characteristics and annotation structure were summarized using counts and proportions, including phenotype prevalence, vote distributions across raters and the proportion of labels affected by disagreement. Because the cohorts differed in size, age range, phenotype composition, annotation structure and video quality, these descriptive summaries were used to contextualize downstream model performance. Inter-rater agreement was quantified separately for each dataset and phenomenology. For datasets rated by two experts, pairwise agreement was summarized using Cohen's kappa. For the dataset rated by five experts, multi-rater agreement was summarized using Fleiss' kappa. Raw agreement proportions were also reported to complement kappa-based measures, particularly for low-prevalence phenomenologies in which chance-corrected agreement coefficients may be unstable so inappropriate. Baseline and calibrated inference results were analyzed using the same metric framework to enable direct comparison before and after local calibration. Given the modest size of the external dataset these analyses were interpreted primarily as descriptive and exploratory rather than confirmatory.

\subsection{Implementation and reproducibility}

All preprocessing, feature extraction, model training, threshold tuning, inference, calibration, and evaluation steps were implemented in Python. The full analysis workflow was organized as a script-based pipeline covering training, patient-level threshold and aggregation tuning, inference, generation of multi-rater ground-truth files, and dataset-specific calibration procedures. To support reproducibility while preserving participant privacy, the derived feature tables used for model development and evaluation is available at \url{https://github.com/xaviervasques/cody-pipeline.git} (data repository). The original clinical videos are not be publicly shared. The complete codebase required to reproduce the preprocessing, modeling, calibration, and evaluation pipeline is available at \url{https://github.com/xaviervasques/cody-pipeline.git}. The complete results were shared on \textbf{Supplementary Material S3}.

\textbf{RESULTS}

\section{Results}

\subsection{Cohorts and phenomenology distribution}

The training cohort comprised 25 participants, 21 patients with combined HMDs and 4 healthy controls, recorded under a standardized CODY-SAMP examination protocol at Beau Soleil Clinic, Montpellier. One independent external dataset was used for inference, without backbone retraining: Pediatric dataset (n=12 children with confirmed monogenic MDs, non-standardized routine clinical videos). Demographic, etiological and acquisition characteristics of the three cohorts are provided in \textbf{Table 2}.

\begin{table}[htbp]\centering\footnotesize
\caption*{\textbf{Table 2.} Demographic, etiological, phenomenological and acquisition characteristics of the training cohort and the external pediatric inference dataset.}
\renewcommand{\arraystretch}{1.15}
\begin{tabularx}{\textwidth}{@{}p{5.2cm}p{4.4cm}X@{}}
\toprule
\textbf{Characteristic} & \textbf{Training cohort} (adult, model training) & \textbf{Pediatric inference cohort} (external, model inference) \\
\midrule
\multicolumn{3}{@{}l}{\textit{Demographics}} \\
Number of patients & 21 & 12 \\
Number of healthy controls & 4 & 0 \\
Sex, female, n (\%) & 12 (57.1\%) & 9 (75.0\%) \\
Age, mean ± SD (years) & 46.9 ± 21.0 & 11.9 ± 2.6 \\
Age, range (years) & 17–75 & 7.0–15.5 \\
\multicolumn{3}{@{}l}{\textit{Etiology}} \\
Monogenic MD & 7 & 12 (all) \\
Acquired or developmental MD & 2 & 0 \\
Unsolved combined MD & 12 & 0 \\
\multicolumn{3}{@{}l}{\textit{Phenomenology composition, n (\%)}\,$\dagger$} \\
Dystonia & 21 (100.0\%) & 12 (100.0\%) \\
Tremor & 13 (61.9\%) & 0 (0.0\%) \\
Myoclonus & 13 (61.9\%) & 4 (33.3\%) \\
Chorea & 4 (19.0\%) & 4 (33.3\%) \\
Athetosis & 7 (33.3\%) & 5 (41.7\%) \\
Ballismus & 2 (9.5\%) & 2 (16.7\%) \\
Stereotypies & 1 (4.8\%) & 1 (8.3\%) \\
Tics & 2 (9.5\%) & 0 (0.0\%) \\
\multicolumn{3}{@{}l}{\textit{Video acquisition}} \\
Recording protocol & Standardized CODY-SAMP examination & Non-standardized; routine pediatric clinical videos \\
Expert raters per patient, n & 2 & 5 \\
\bottomrule
\end{tabularx}

{\footnotesize $\dagger$ Training cohort: per-patient positivity = at least one annotated 10-s window positive; consensus = both raters (DD and LC) agree. Pediatric cohort: consensus = symptom voted positive by ≥3 of 5 raters. MD = movement disorders. CODY-SAMP = standardized clinical video acquisition protocol for combined hyperkinetic MDs.}
\end{table}

In the training cohort, two complementary descriptions of phenomenology composition were obtained.~At the patient level, each participant was characterized by the set of phenomenologies present clinically: dystonia in 21/21 patients (100\%), myoclonus and tremor in 13/21 each (62\%), athetosis in 7/21 (33\%), chorea in 4/21 (19\%), ballismus and tics in 2/21 each (10\%), and stereotypies in 1/21 (5\%).~At the window level, the same patients yielded a different distribution because clinicians annotated only those 10-s windows in which a phenomenology was actually visible, rather than every window of a positive patient. Window-level positivity was therefore lower and more uneven, with dystonia dominant, tremor and myoclonus well represented, and ballismus, stereotypies and tics each contributing only a small fraction of the annotated windows. As a consequence, the supervised learning signal available to the predictive backbone was substantially richer for the four dominant phenomenologies than for the three rare ones.

In the pediatric dataset, the proportion of labelled-present phenomenologies under the main present/absent definition was 29.2\% (28 of 96). The dataset was dominated by dystonia (12/12 patients), followed by athetosis (5/12), chorea (4/12) and myoclonus (4/12), with tremor and tics absent (\textbf{Table 2}, \textbf{Figure 2}).

This composition has direct methodological implications. In the pediatric dataset, three of the eight labels are invariant (dystonia is universally present; tremor and tics are uniformly absent). Performance metrics computed over all eight labels therefore carry contributions from columns with zero variance, which can inflate Hamming accuracy through trivially-correct true negatives. The pediatric external dataset was rated independently by five expert clinicians (LC, DD, GH, MCJ, JDOE), each scoring the presence or absence of each of the eight target hyperkinetic phenomenologies at the patient level, based on the standard clinical definitions of each phenomenology as detailed in the Methods section. Patient-level present-versus-absent labels could therefore be derived under several consensus rules (\textbf{Figure 2}). Under the main definition, a phenomenology was considered present at the patient level when voted positive by at least three of the five raters (Majority ≥3/5); this yielded 29.2\% of the 96 patient--symptom labels classified as positive. Gradually increasing the agreement requirement progressively reduced the proportion of labels considered present, to 22.9\% under Majority (≥4/5) and 18.8\% under unanimous agreement (5/5). A second, complementary definition restricted the evaluation to labels with unambiguous expert agreement on both sides i.e., the symptom voted positive by ≥3/5 (or ≥4/5, or 5/5) raters and simultaneously voted negative by 0/5 raters on the opposite side. Under this restricted definition, between 78.1\% and 67.7\% of patient--symptom labels remained evaluable as agreement requirements increased. Together, these analyses show that both the apparent phenotype frequency and the size of the evaluable clinical label space depended on how strictly expert agreement was defined.

\begin{figure}[htbp]
\centering
\includegraphics[width=\textwidth,keepaspectratio]{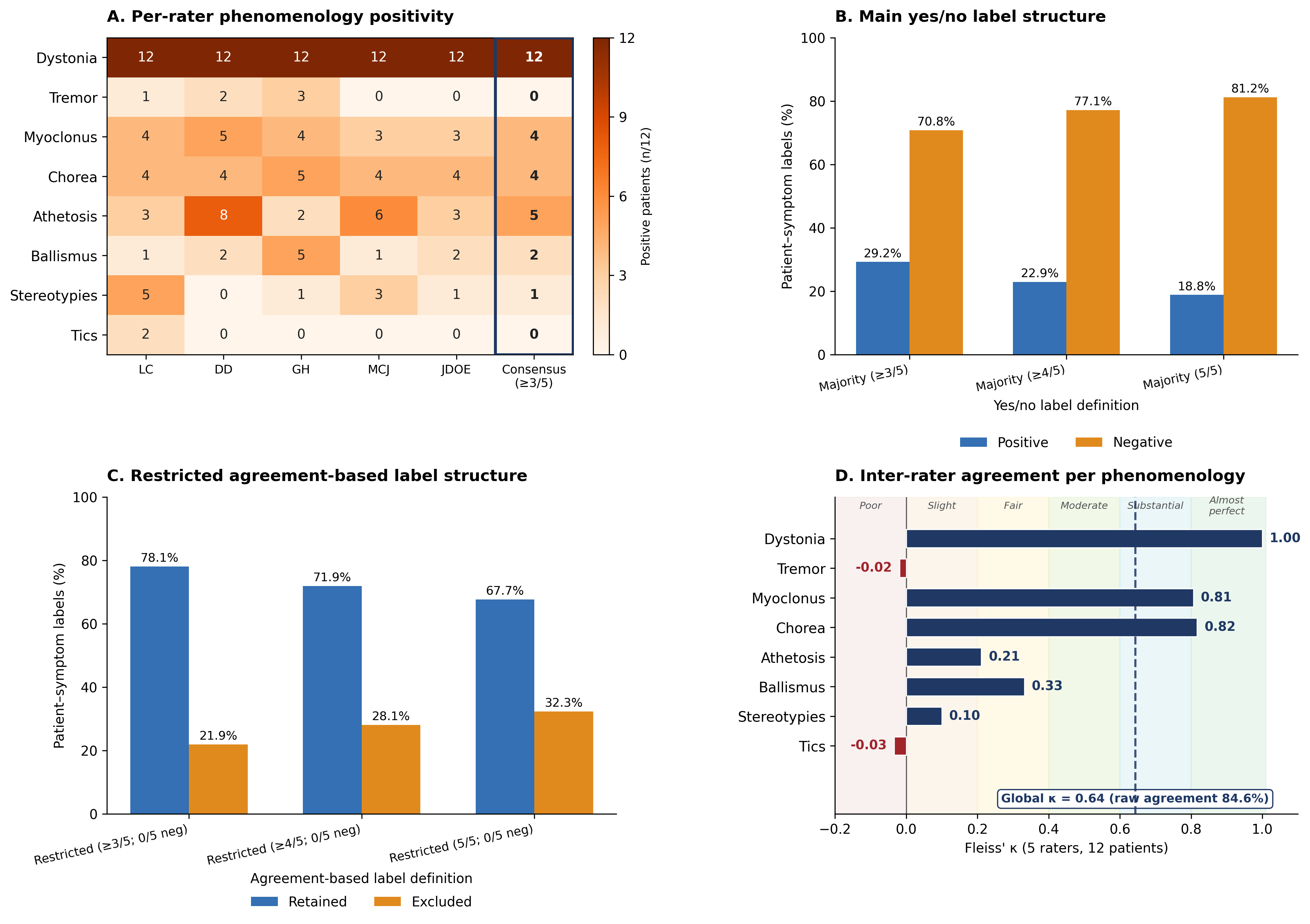}
\caption*{\textbf{Figure 2.} (A)~The eight target phenomenologies were rated by five expert clinicians (LC, DD, GH, MCJ, JDOE) for each of the 12 pediatric patients (96 patient--symptom labels in total), and a patient-level consensus was derived as the symptom being voted positive by at least three of the five raters. Dystonia was present in all 12 patients with unanimous agreement, athetosis in 5 patients, myoclonus and chorea in 4 patients each, ballismus in 2, stereotypies in 1, while neither tremor nor tics reached consensus positivity.~(D)~Overall inter-rater agreement was substantial (Fleiss\textquotesingle{} κ = 0.64; raw agreement 84.6\%) but varied markedly across phenomenologies, being almost perfect for dystonia, substantial for chorea and myoclonus, fair for athetosis and ballismus, and slight or absent for stereotypies, tremor and tics.~(B--C)~The proportion of patient--symptom labels classified positive depended on the consensus definition.}
\end{figure}

\subsection{Inter-rater agreement and annotation ambiguity}

Overall inter-rater agreement in the pediatric dataset was substantial: across the 96 patient--symptom labels, pairwise raw agreement between raters averaged 84.6\% and Fleiss\textquotesingle{} κ reached 0.64. Agreement varied substantially by phenomenology (\textbf{Figure 2D}). It was substantial for chorea and myoclonus (κ = 0.82 and 0.81; raw agreement 91.7\% for both), fair for ballismus (κ = 0.33; raw 80.0\%) and athetosis (κ = 0.21; raw 63.3\%, the lowest of all phenomenologies), only slight for stereotypies (κ = 0.10), and near zero for tremor (κ = −0.02) and tics (κ = −0.03). The latter two values reflect the very low prevalence of these phenomenologies in the cohort, only a few isolated positive votes were cast across raters, a regime in which κ is known to be unstable and largely uninformative, despite raw agreement remaining moderate to high (81.7\% and 93.3\% respectively). Dystonia was uniformly labelled present by all five raters in every patient, yielding raw agreement of 100\%; κ is formally undefined for such an invariant label and is shown as 1.00 by convention in \textbf{Figure 2D}.

Annotation uncertainty was therefore not uniform across phenomenologies but concentrated in specific labels, with the strongest inter-rater divergence on athetosis (range 2--8 positive patients across raters, \textbf{Figure 2A}), stereotypies (range 0--5) and ballismus (range 1--5). This pattern motivated the parallel reporting of model performance under the main present/absent definition and under analyses restricted to labels with unambiguous expert agreement (referred to throughout as the restrictive agreement-based evaluation).

\subsection{Patient-level performance before and after local calibration}

Patient-level performance was compared between the baseline (uncalibrated) and locally calibrated deployments using Hamming accuracy and the Jaccard index, under the main present/absent and restrictive agreement-based definitions, each at three rater-agreement levels (≥3/5, ≥4/5, 5/5). Unless stated otherwise, results refer to the seven held-out patients. Results are provided in \textbf{Table 3}, with the corresponding metric and confusion-structure changes shown in \textbf{Figures 3 and 4}.

\begin{table}[htbp]\centering\scriptsize
\caption*{\textbf{Table 3.} Patient-level performance and aggregated confusion structure before and after local calibration in the external pediatric cohort. Baseline = uncalibrated decision layer (threshold 0.5, p95 aggregation); Calibrated = per-label thresholds and aggregation fitted on the five clinician-selected calibration patients. Metrics are patient-averaged; TP/TN/FP/FN are aggregated counts pooled across patients.}
\renewcommand{\arraystretch}{1.1}\setlength{\tabcolsep}{4pt}
\begin{tabular}{@{}lllcccccc@{}}
\toprule
\textbf{Evaluation} & \textbf{Agr.} & \textbf{Deployment} & \textbf{TP} & \textbf{TN} & \textbf{FP} & \textbf{FN} & \textbf{Hamming} & \textbf{Jaccard} \\
\midrule
\multicolumn{9}{@{}l}{\textit{Held-out (P1,P2,P4,P6,P7,P9,P12; n=7) — external validation}} \\
\midrule
\multirow{6}{*}{Main present/absent}
 & \multirow{2}{*}{≥3/5} & Baseline & 8 & 37 & 2 & 9 & 0.804 & 0.548 \\
 &  & Calibrated & 13 & 34 & 5 & 4 & 0.839 & 0.633 \\
 & \multirow{2}{*}{≥4/5} & Baseline & 7 & 39 & 3 & 7 & 0.821 & 0.536 \\
 &  & Calibrated & 12 & 36 & 6 & 2 & 0.857 & 0.655 \\
 & \multirow{2}{*}{5/5} & Baseline & 7 & 42 & 3 & 4 & 0.875 & 0.655 \\
 &  & Calibrated & 9 & 36 & 9 & 2 & 0.804 & 0.512 \\
\midrule
\multirow{6}{*}{Restrictive agreement-based}
 & \multirow{2}{*}{≥3/5} & Baseline & 8 & 24 & 2 & 9 & 0.735 & 0.548 \\
 &  & Calibrated & 13 & 24 & 2 & 4 & 0.856 & 0.714 \\
 & \multirow{2}{*}{≥4/5} & Baseline & 7 & 24 & 2 & 7 & 0.762 & 0.560 \\
 &  & Calibrated & 12 & 24 & 2 & 2 & 0.900 & 0.786 \\
 & \multirow{2}{*}{5/5} & Baseline & 7 & 24 & 2 & 4 & 0.829 & 0.679 \\
 &  & Calibrated & 9 & 24 & 2 & 2 & 0.893 & 0.762 \\

\midrule
\multicolumn{9}{@{}l}{\textit{All patients (n=12)}} \\
\midrule
\multirow{6}{*}{Main present/absent}
 & \multirow{2}{*}{≥3/5} & Baseline & 13 & 65 & 3 & 15 & 0.812 & 0.507 \\
 &  & Calibrated & 21 & 61 & 7 & 7 & 0.854 & 0.654 \\
 & \multirow{2}{*}{≥4/5} & Baseline & 12 & 70 & 4 & 10 & 0.854 & 0.590 \\
 &  & Calibrated & 18 & 64 & 10 & 4 & 0.854 & 0.618 \\
 & \multirow{2}{*}{5/5} & Baseline & 12 & 74 & 4 & 6 & 0.896 & 0.674 \\
 &  & Calibrated & 15 & 65 & 13 & 3 & 0.833 & 0.549 \\
\midrule
\multirow{6}{*}{Restrictive agreement-based}
 & \multirow{2}{*}{≥3/5} & Baseline & 13 & 44 & 3 & 15 & 0.752 & 0.507 \\
 &  & Calibrated & 21 & 43 & 4 & 7 & 0.850 & 0.701 \\
 & \multirow{2}{*}{≥4/5} & Baseline & 12 & 44 & 3 & 10 & 0.802 & 0.604 \\
 &  & Calibrated & 18 & 43 & 4 & 4 & 0.885 & 0.750 \\
 & \multirow{2}{*}{5/5} & Baseline & 12 & 44 & 3 & 6 & 0.853 & 0.688 \\
 &  & Calibrated & 15 & 43 & 4 & 3 & 0.897 & 0.764 \\

\midrule
\multicolumn{9}{@{}l}{\textit{In-sample calibration set (P3,P5,P8,P10,P11; n=5) — for completeness only}} \\
\midrule
\multirow{6}{*}{Main present/absent}
 & \multirow{2}{*}{≥3/5} & Baseline & 5 & 28 & 1 & 6 & 0.825 & 0.450 \\
 &  & Calibrated & 8 & 27 & 2 & 3 & 0.875 & 0.683 \\
 & \multirow{2}{*}{≥4/5} & Baseline & 5 & 31 & 1 & 3 & 0.900 & 0.667 \\
 &  & Calibrated & 6 & 28 & 4 & 2 & 0.850 & 0.567 \\
 & \multirow{2}{*}{5/5} & Baseline & 5 & 32 & 1 & 2 & 0.925 & 0.700 \\
 &  & Calibrated & 6 & 29 & 4 & 1 & 0.875 & 0.600 \\
\midrule
\multirow{6}{*}{Restrictive agreement-based}
 & \multirow{2}{*}{≥3/5} & Baseline & 5 & 20 & 1 & 6 & 0.776 & 0.450 \\
 &  & Calibrated & 8 & 19 & 2 & 3 & 0.843 & 0.683 \\
 & \multirow{2}{*}{≥4/5} & Baseline & 5 & 20 & 1 & 3 & 0.858 & 0.667 \\
 &  & Calibrated & 6 & 19 & 2 & 2 & 0.863 & 0.700 \\
 & \multirow{2}{*}{5/5} & Baseline & 5 & 20 & 1 & 2 & 0.888 & 0.700 \\
 &  & Calibrated & 6 & 19 & 2 & 1 & 0.903 & 0.767 \\

\bottomrule
\end{tabular}
\end{table}

Before calibration, under the main present/absent definition at the principal ≥3/5 level, the backbone already conveyed clinically meaningful signal but did not reconstruct complete phenotype profiles: Hamming accuracy was 0.804 while the Jaccard index was lower, at 0.548. This pattern, relatively high Hamming, lower Jaccard, is clinically informative, indicating that the backbone identified absent phenomenologies (true negatives, which dominate by prevalence) more readily than it recovered the present phenomenologies that carry clinical meaning.

After local calibration of the decision step, performance improved on the held-out patients. Under the main ≥3/5 definition, Hamming accuracy rose from 0.804 to 0.839 and the Jaccard index from 0.548 to 0.633; the confusion structure shifted accordingly, with true positives increasing from 8 to 13 and false negatives falling from 9 to 4, at the cost of a modest rise in false positives (2 to 5). Under the restrictive agreement-based definition at the same level, the gain was larger and came without that cost: Hamming accuracy rose from 0.735 to 0.856 and the Jaccard index from 0.548 to 0.714, with true positives increasing from 8 to 13, false negatives falling from 9 to 4, and false positives unchanged (\textbf{Table 4}; \textbf{Figure 4}). Local calibration therefore improved sensitivity to present phenomenologies while preserving control of false positives, most clearly when low-agreement labels were not forced into a binary present/absent decision.

Because the calibration subset and the held-out set were disjoint, we could confirm that these gains were not an artefact of overfitting. Performance on the five calibration patients was close to that on the seven held-out patients (calibrated Jaccard 0.683 versus 0.633 under the main ≥3/5 definition), indicating that a decision layer tuned on a few representative patients generalised to unseen patients from the same cohort.

Post-calibration performance nonetheless depended on how the reference labels were defined (\textbf{Figure 3}). Under the restrictive agreement-based definition, the calibrated advantage was consistent and broadly increased as the agreement requirement was tightened (Jaccard 0.714, 0.786 and 0.762 at ≥3/5, ≥4/5 and 5/5). Under the main present/absent definition, the advantage was present at the standard ≥3/5 level but narrowed and then reversed at the strictest level, where the calibrated Jaccard fell below baseline (0.512 versus 0.655 at 5/5). The reason is visible in the confusion structure (\textbf{Figure 4}): the lower per-phenotype thresholds introduced by calibration add a small number of false positives, and a strict present/absent reference penalises these more heavily than a definition that masks low-agreement labels. The clinical reading is that apparent performance reflects, in part, the stringency and construction of the reference standard, and that the benefit of calibration is most robust when evaluation focuses on phenomenologies with more definite expert agreement.

\begin{table}[htbp]\centering\footnotesize
\caption*{\textbf{Table 4.} Per-phenotype confusion structure before and after local calibration, under the main present/absent definition (≥3/5). Counts are reported for the seven held-out patients and for all twelve patients of the external pediatric cohort.}
\renewcommand{\arraystretch}{1.1}\setlength{\tabcolsep}{5pt}
\begin{tabular}{@{}ll cccc cccc@{}}
\toprule
& & \multicolumn{4}{c}{\textbf{Held-out (n=7)}} & \multicolumn{4}{c}{\textbf{All patients (n=12)}} \\
\cmidrule(lr){3-6}\cmidrule(lr){7-10}
\textbf{Phenotype} & \textbf{Deployment} & TP & TN & FP & FN & TP & TN & FP & FN \\
\midrule
\multirow{2}{*}{Dystonia} & Baseline & 7 & 0 & 0 & 0 & 12 & 0 & 0 & 0 \\
 & Calibrated & 7 & 0 & 0 & 0 & 12 & 0 & 0 & 0 \\
\addlinespace[2pt]
\multirow{2}{*}{Tremor} & Baseline & 0 & 7 & 0 & 0 & 0 & 12 & 0 & 0 \\
 & Calibrated & 0 & 7 & 0 & 0 & 0 & 12 & 0 & 0 \\
\addlinespace[2pt]
\multirow{2}{*}{Myoclonus} & Baseline & 0 & 4 & 0 & 3 & 0 & 8 & 0 & 4 \\
 & Calibrated & 0 & 4 & 0 & 3 & 0 & 8 & 0 & 4 \\
\addlinespace[2pt]
\multirow{2}{*}{Chorea} & Baseline & 0 & 3 & 1 & 3 & 0 & 7 & 1 & 4 \\
 & Calibrated & 3 & 2 & 2 & 0 & 4 & 5 & 3 & 0 \\
\addlinespace[2pt]
\multirow{2}{*}{Athetosis} & Baseline & 0 & 5 & 0 & 2 & 0 & 7 & 0 & 5 \\
 & Calibrated & 2 & 3 & 2 & 0 & 4 & 4 & 3 & 1 \\
\addlinespace[2pt]
\multirow{2}{*}{Ballismus} & Baseline & 1 & 5 & 1 & 0 & 1 & 8 & 2 & 1 \\
 & Calibrated & 1 & 5 & 1 & 0 & 1 & 9 & 1 & 1 \\
\addlinespace[2pt]
\multirow{2}{*}{Stereotypies} & Baseline & 0 & 6 & 0 & 1 & 0 & 11 & 0 & 1 \\
 & Calibrated & 0 & 6 & 0 & 1 & 0 & 11 & 0 & 1 \\
\addlinespace[2pt]
\multirow{2}{*}{Tics} & Baseline & 0 & 7 & 0 & 0 & 0 & 12 & 0 & 0 \\
 & Calibrated & 0 & 7 & 0 & 0 & 0 & 12 & 0 & 0 \\
\addlinespace[2pt]
\bottomrule
\end{tabular}
\end{table}

\begin{figure}[htbp]
\centering
\includegraphics[width=\textwidth,keepaspectratio]{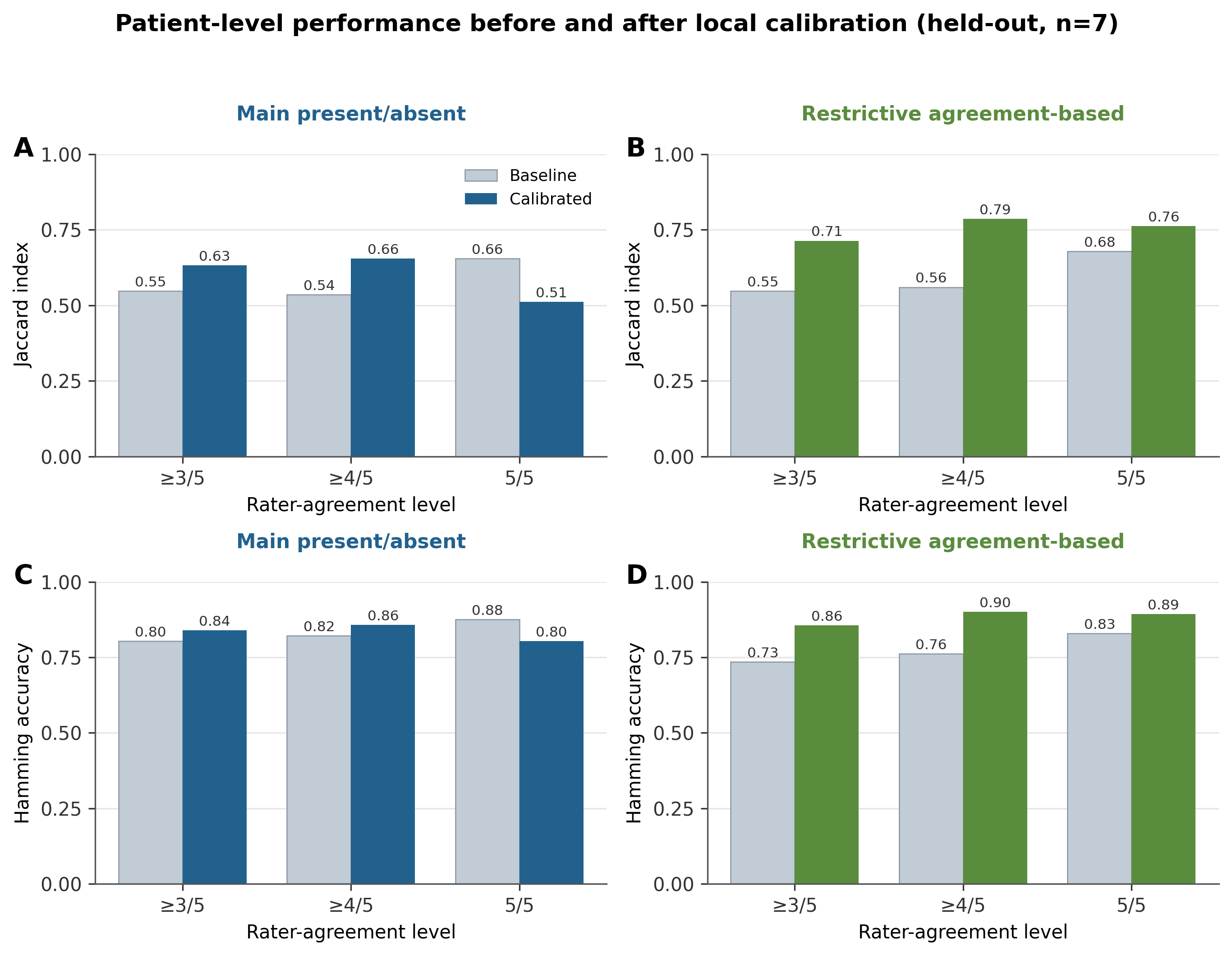}
\caption*{\textbf{Figure 3.}~Patient-level performance before and after local calibration (held-out cohort).~Jaccard index (A, B) and Hamming accuracy (C, D) for the baseline (uncalibrated) and locally calibrated deployments on the seven held-out pediatric patients, under the main present/absent definition (A, C) and the restrictive agreement-based definition (B, D), at three rater-agreement levels (≥3/5, ≥4/5, 5/5). Under the restrictive agreement-based definition the calibrated advantage is consistent across levels, whereas under the main present/absent definition it narrows and reverses at the strictest level, reflecting the influence of the reference-label construction.}
\end{figure}

\begin{figure}[htbp]
\centering
\includegraphics[width=\textwidth,keepaspectratio]{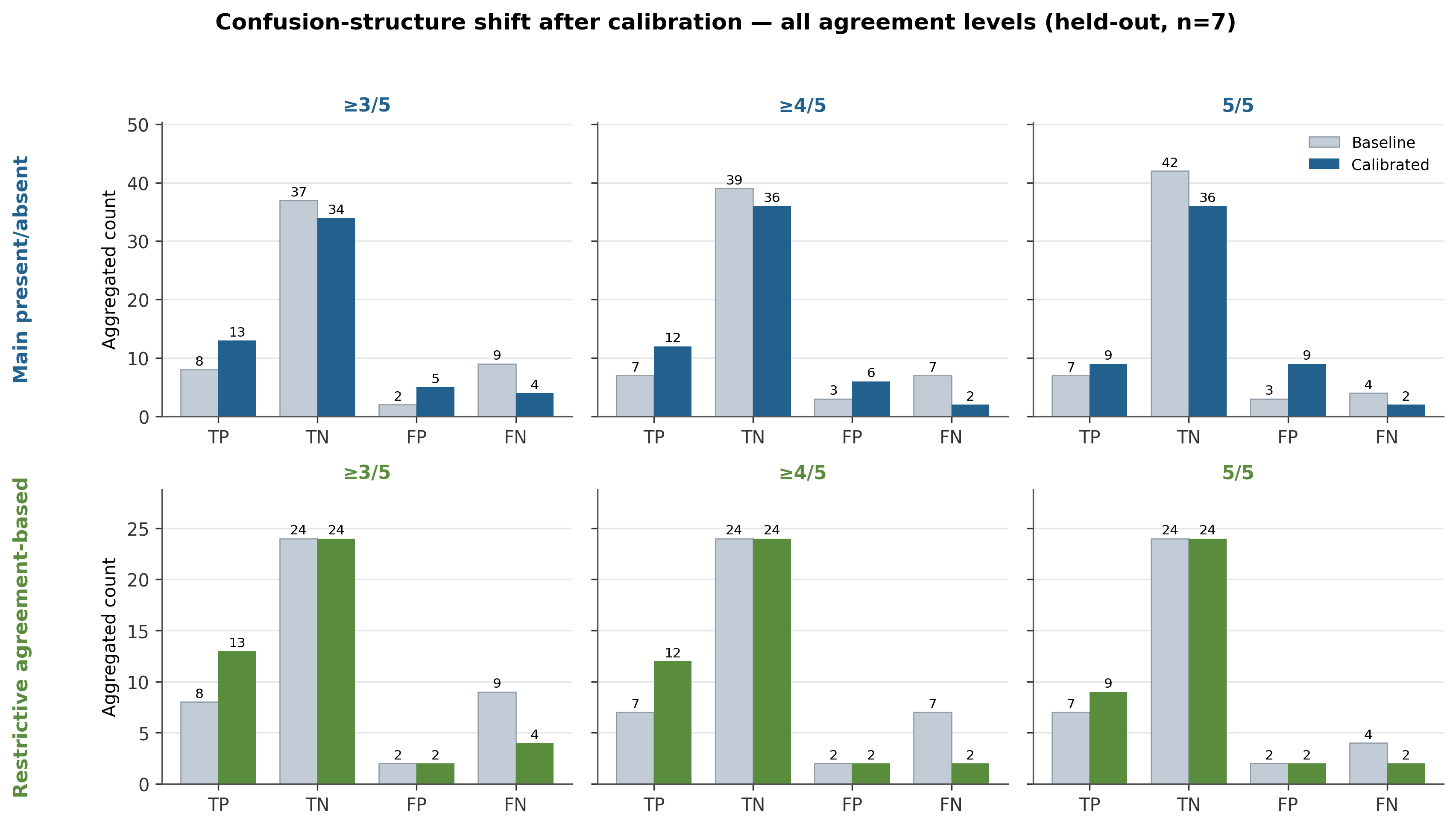}
\caption*{\textbf{Figure 4.} Confusion-structure shift after local calibration (held-out cohort).~Aggregated true positives (TP), true negatives (TN), false positives (FP) and false negatives (FN) for the baseline and locally calibrated deployments on the seven held-out patients. Top row, main present/absent definition; bottom row, restrictive agreement-based definition; columns correspond to rater-agreement levels ≥3/5, ≥4/5 and 5/5. After calibration, true positives increase and false negatives fall in both families; under the main present/absent definition this is accompanied by a modest rise in false positives that grows at stricter agreement levels, whereas false positives remain stable under the restrictive agreement-based definition.}
\end{figure}

\subsection{Phenotype-specific analyses}

The effect of calibration differed markedly across phenomenologies, and the pattern was clinically informative in its own right (\textbf{Table 4}). Three response profiles emerged: phenomenologies already detected by the backbone, phenomenologies recovered by calibration, and phenomenologies that remained undetected despite calibration.

Dystonia, present in every patient and the dominant phenomenology of the cohort, was detected reliably both before and after calibration (7 of 7 held-out patients; sensitivity and specificity of 1.0). Its detection did not depend on calibration: the backbone assigned consistently high per-window probabilities, so dystonia was captured even under the generic baseline decision rule, and calibration only confirmed this. The shared backbone had thus robustly encoded the cohort\textquotesingle s most prevalent phenotype.

Chorea and athetosis illustrate genuine recovery by calibration. Before calibration both were entirely missed (chorea: 0 true positives, 3 false negatives; athetosis: 0 true positives, 2 false negatives). After calibration, all present cases were recovered (chorea 3 of 3, athetosis 2 of 2; sensitivity 1.0), each at the cost of two false positives. A sensitivity of this kind, achieved with limited overcalling, is clinically meaningful in a decision-support setting: it flags patients who warrant closer specialist review, recovering phenomenologies that the backbone had encoded but the uncalibrated decision rule had suppressed.

By contrast, myoclonus was not recovered under any definition (0 true positives, 3 false negatives, before and after calibration). Since a miscalibrated threshold can be corrected only when the backbone has encoded discriminative signal in the first place, this points to a representational limitation rather than a decision-layer mismatch, consistent with too few myoclonus-positive patients available for the adult backbone to learn from. Ballismus showed a rare and unstable pattern (a single positive case, detected, with one false positive), constrained both by its rarity and by modest inter-rater agreement (κ = 0.332), which limits what any calibration can achieve.

Finally, tremor and tics had no positive cases in the held-out pediatric cohort, so for these the relevant criterion is specificity rather than sensitivity; calibration preserved correct true-negative assignment without introducing false positives, and the single stereotypies-positive case was not recovered. Taken together, these analyses show that local calibration serves a dual clinical function: recovering missed positive phenomenologies where backbone signal is present, and protecting against overcalling of absent ones, both of which matter for deployment in routine care.

\section{Discussion}

In this study, we developed and evaluated a video-based framework for patient-level phenotyping of combined hyperkinetic MDs from routine clinical recordings. The central result is that clinically meaningful phenotype information could be transferred across external cohorts using a shared predictive backbone based on markerless pose estimation, interpretable kinematic descriptors and a pretrained tabular foundation model, while substantial performance gains were achieved by locally recalibrating only the final patient-level decision step. This two-stage design was chosen because, from a clinical implementation perspective, it is substantially lighter than full redevelopment of the model, while still allowing adaptation to local cohort characteristics such as phenotype prevalence, annotation style and video conditions.~Importantly, the magnitude of this benefit was not uniform, but depended on the amount of informative data available for a given phenomenology, the degree of expert agreement, and the clinical and technical characteristics of the target dataset. This finding is important because it suggests a practical strategy for digital phenotyping in MDs: rather than retraining a full model for every new cohort, one may preserve a common representation and prediction backbone while adapting the final aggregation and thresholding step to local conditions. In the present study, this strategy yielded consistent gains and was particularly effective in a heterogeneous, routine pediatric cohort, where local calibration improved both recovery of clinically relevant positive labels and control of false positives.~More broadly, these results argue for stronger data sharing and harmonization across centers, because robust digital phenotyping of rare and heterogeneous MDs will likely require larger pooled cohorts spanning wider age ranges, recording conditions and expert annotation profiles.

The etiological heterogeneity was deliberate in our study: first, it is the population targeted by the comprehensive CODY-SAMP clinical scale we developed and second, by training on a broad phenotypic range, we aimed to maximize backbone generalizability across distinct motor phenomenologies rather than optimizing for a single disorder. The age ranges of the one main datset did not overlap (training cohort: 17--75 years ; pediatric dataset\_2: 7.0--15.5 years), so that the pediatric and adult external inference datasets represented genuinely distinct demographic regimes relative to the adult development cohort. As it relates to the acquisition conditions, the study aimed not only to evaluate performance in standardized recordings as in the development cohort, but also to test whether the framework could remain informative when applied to suboptimal and low quality real-world clinical videos. The YOLOv8-based pose-estimation approach was selected over alternatives (such as MediaPipe or OpenPose) because of its robustness to partial occlusion, its validated performance on clinical video material across body regions, and its compatibility with variable video resolutions typical of routine care recordings.

Our findings align with the broader evolution of digital neurology, in which video-based recognition and quantification is increasingly viewed as a scalable and non-invasive complement to bedside examination rather than a replacement for clinical expertise \textsuperscript{1,3--5,7}. Recent reviews have emphasized both the promise and the immaturity of the field: while computer vision and video analysis are rapidly expanding in MDs , many published systems remain focused on single phenomenology, narrowly controlled tasks or highly specific acquisition conditions \textsuperscript{6,7,14}. By contrast, the present framework addressed a harder and more clinically relevant problem, namely the recognition of multiple co-occurring hyperkinetic phenomenologies from heterogeneous clinical practice videos at the patient level. This multi-symptom setting better reflects real clinical practice, where mixed phenomenology is common and where diagnostic reasoning depends less on a single motor sign than on the structured combination of several abnormal movements \textsuperscript{3,4}. The fact that useful signal could be extracted from ordinary videos and translated into patient-level phenotype profiles therefore supports the broader idea that clinically deployed digital phenotyping can be built from routine audiovisual material, provided that the computational pipeline remains robust to annotation uncertainty and cross-site variability.

A major strength of the method is the deliberate separation between a shared predictive backbone and a dataset-adaptive decision layer. Beyond the technical mechanics described in the Methods section, this design choice has direct implications for the realistic adoption of digital phenotyping in routine clinical care. Many published video-AI systems require centralized retraining or substantial dataset-specific re-engineering to be transferred to new sites, which makes deployment in resource-constrained pediatric centers difficult. By contrast, the present framework requires only that a small labeled subset already exist on the target site, a condition that is naturally met when expert phenotyping is part of standard clinical practice. Conceptually, this is appealing because it reflects a realistic form of clinical transfer learning. The pose-estimation and feature-extraction stages aim to encode movement in a relatively site-agnostic way, while the patient-level aggregation and thresholding stages are allowed to absorb local differences in prevalence, annotation culture, video quality and recording context. In other words, the framework does not assume that external failure necessarily reflects failure of representation learning. Instead, it treats part of external performance degradation as a decision-calibration problem. This is particularly relevant in clinical AI, where domain shift can arise from many factors beyond the biological signal itself, including acquisition conditions, case mix, annotation framework and prevalence structure \textsuperscript{15,16}. The meaningful improvement observed after local recalibration suggests that this decomposition was not only conceptually elegant but empirically useful in practice. This pattern is conceptually important and applies more broadly to clinical AI in heterogeneous populations. A predictive backbone that "fails" before calibration in the sense of yielding zero true positives is not necessarily a backbone that fails to encode the target phenomenon. Two distinct failure modes must be distinguished: (i) representational failure, where the backbone has not learned a feature space in which positive and negative cases can be separated, in which case no calibration of the final decision step can rescue performance; and (ii) decision-layer mismatch, where the backbone produces an informative score distribution but the threshold tuned in the development cohort is misaligned with the prevalence and probability scale of the target cohort. The first situation requires retraining; the second only requires adjusting how the existing scores are turned into present/absent decisions.The improvements observed in this study, in which calibration restored sensitivity without altering the backbone, are consistent with the second mechanism and argue against discarding the underlying model on the basis of uncalibrated external performance alone.

Another important strength lies in the choice to use interpretable kinematic descriptors as previously published \textsuperscript{5} rather than an entirely end-to-end black-box video classifier. Markerless pose estimation has become an increasingly attractive approach in MDs research because it allows standard videos to be transformed into anatomically meaningful trajectories without dedicated motion-capture hardware \textsuperscript{7,14}. Building structured kinematic descriptors on top of these trajectories preserves an interpretable link between raw video and downstream decision-making. This is advantageous for at least three reasons. First, it reduces dependence on highly standardized visual appearance and may improve portability across cameras and environments. Second, it creates an explicit bridge to clinical motor phenomenology by summarizing movement amplitude, variability, rhythmicity, spectral structure and complexity. Third, it yields structured tabular inputs that can be reused, audited and reanalyzed independently of the original video stream. In translational settings, this form of interpretability is likely to matter for clinician trust, debugging and future regulatory evaluation \textsuperscript{8}.

The use of a pretrained tabular foundation model is also a key methodological contribution of this work. Once pose-derived movement signals are converted into kinematic descriptors, the learning problem becomes tabular rather than purely visual. Classical approaches such as logistic regression, support vector machines, random forests or gradient-boosted trees remain strong baselines in tabular learning, but they typically require repeated model selection and extensive dataset-specific tuning, especially when datasets are small, imbalanced and heterogeneous \textsuperscript{5}. Foundation models for tabular data propose a different solution: they amortize statistical learning across a very large distribution of synthetic tasks and then transfer this inductive bias to new real-world datasets \textsuperscript{9,10,17}. Recent work has shown that such models can achieve state-of-the-art performance on small to medium-sized tabular tasks, often outperforming heavily tuned conventional baselines \textsuperscript{9,10,17}. In the context of rare or heterogeneous clinical datasets, this is especially attractive because it shifts part of the modeling burden away from local data scarcity toward transferable prior knowledge encoded during pretraining.

In our study, the practical value of this choice is not simply that a foundation model was used, but that it enabled a deployment logic centered on reuse and generalizability. The shared predictive backbone already transferred a measurable signal to the external pediatric cohort before local adaptation, indicating that the pretrained model did not merely memorize the development cohort. The subsequent performance gains after recalibration further suggest that the main residual mismatch lay, at least in part, in the final decision layer rather than in the entire representation space. This is precisely the type of behavior one would hope for from a clinically useful foundation model: robust enough to generalize non-trivially beyond the original site, yet flexible enough to benefit from light-touch local adaptation. More broadly, this supports the view that foundation models may be particularly valuable in clinical tabular settings where annotated sample sizes remain modest and where retraining bespoke models at every new institution is unrealistic \textsuperscript{9,10,17}.

At the same time, the results also show that the method should not be interpreted as uniformly successful across all phenotypes. The clearest calibration gains were concentrated in chorea and athetosis, both recovered from being undetected at baseline, whereas dystonia, the dominant phenomenology of the cohort, was already detected before calibration; myoclonus, by contrast, remained undetected, and tremor, tics and stereotypies had too few or no positive cases in this cohort to be meaningfully evaluated. This phenotype dependence is clinically plausible. Different hyperkinetic phenomenologies vary in temporal structure, frequency content, anatomical expression and inter-rater consistency, and some may be intrinsically harder to capture with 2D pose-derived descriptors alone. In addition, low-prevalence labels are affected by both statistical scarcity and unstable agreement structure, making reliable supervised learning and evaluation more difficult. The present findings therefore argue against a simplistic reading of ``video AI for MDs'' as a uniform task. Rather, they support the view that digital phenotyping performance will remain phenotype-specific and conditioned by both the signal characteristics of the movement and the quality and quantity of the reference labels.

A further strength of the study is that it did not collapse multi-rater annotations into a single unquestioned ground truth. Instead, the evaluation explicitly retained complementary binary and agreement-aware reference schemes. This is particularly important in MDs, where phenomenological boundaries can be uncertain and inter-rater disagreement is not merely noise but part of the clinical reality \textsuperscript{3,4}. By reporting performance under several reference-label constructions, the study makes visible the extent to which apparent model quality depends on the definition of the target itself. This is an intellectually honest and clinically valuable choice. It also helps explain why some gains were strongest under the restrictive agreement-based evaluation: not all apparent improvement reflects the same underlying task difficulty, because some evaluation schemes concentrate on more consensual cases while others preserve more ambiguity. In digital medicine, where algorithmic performance is often reported against a deceptively fixed reference standard, such explicit treatment of annotation uncertainty should be seen as a methodological advantage rather than a complication \textsuperscript{8}.

\subsection{Limitations}

Several limitations should be acknowledged. First, the study is constrained by cohort size: external pediatric dataset\_2 (n=12) is small, and performance estimates, particularly for low-prevalence phenomenologies, should be interpreted as exploratory rather than confirmatory. Larger multi-center cohorts will be required to provide stable performance estimates and to support learning for rare labels such as ballismus and stereotypies. Second, the external cohorts differed not only in patient population but in annotation structure, number of raters, and reference-label construction. This heterogeneity is clinically realistic and scientifically informative, but complicates direct cross-cohort comparison. The decision to evaluate under multiple label definitions is a methodological strength, but also adds analytic complexity that may limit accessibility for readers. Third, the pediatric dataset represented a particularly adverse acquisition setting, pediatric routine videos with non-fixed cameras, variable framing, and overall poorer visual quality. The fact that calibration still yielded substantial gains in this cohort is encouraging, but performance should also be interpreted through the lens of acquisition difficulty. Video quality affected keypoint detection reliability, introduced missing trajectories, and likely degraded posture-derived features, all of which constrained the signal available to the backbone. Fourth, the framework relies on 2D pose trajectories from monocular video, which imposes representational limits for phenomenologies dependent on depth, distal fine movements, or muscle overflow, limitations that likely contributed to the persistent detection failure for myoclonus. Three-dimensional tracking and higher-frame-rate acquisition are natural candidates for future improvement. Fifth, the study did not include a head-to-head benchmarking of the tabular foundation model against conventional classifiers within the same feature space. The conceptual and empirical rationale for the foundation-model approach is well supported by the literature,\textsuperscript{8,9,10,17} but explicit comparison against tuned random forests, gradient-boosted trees, or support vector machines in the same experimental configuration would strengthen the methodological contribution of this work.

\subsection{Future directions}

Several directions emerge naturally from this work. At the scale level, larger multi-center datasets spanning broader age ranges, etiologies, and recording conditions are needed to estimate generalization reliably and to build stable learning support for rare phenotypes. Data sharing initiatives and harmonized annotation protocols, ideally using CODY-SAMP or a compatible framework, across specialist centers would substantially accelerate this work.

At the representation level, richer visual encoding should be explored: improved keypoint models with uncertainty estimation, 3D reconstruction from multi-view or depth-sensing setups, and hybrid architectures combining interpretable kinematic features with learned temporal embeddings may improve detection of subtle or intermittent phenomenologies such as myoclonus and athetosis. For DBS patients specifically, the interaction between stimulation state and kinematic signal should be explicitly modeled in future studies, given that DBS-induced clinical changes (improvement or worsening) directly modifies the observable movement phenotype.

At the decision-support level, a natural next step is to expose the patient-level probabilities as continuous confidence indicators rather than binary yes/no outputs, calibrated against expert-derived probabilities via Platt scaling or isotonic regression on the local calibration cohort. Introducing an explicit abstention zone for intermediate scores, consistent with how movement-disorder specialists reason in ambiguous cases, would align the framework with trustworthy AI principles and is consistent with emerging regulatory guidance on clinical AI systems.

Finally, prospective clinical utility studies are needed to determine whether video-based phenotype suggestions improve inter-rater consistency, accelerate referral decisions, reduce time-to-diagnosis in pediatric practice, or support longitudinal monitoring of DBS programming response. These questions cannot be answered by retrospective label agreement alone.

\section{Conclusion}

This study establishes that patient-level phenotyping of combined HMDs, including eight potential phenomenologies assessed simultaneously, is achievable from routine clinical videos, markerless pose estimation, interpretable kinematic descriptors and a pretrained tabular foundation model. The framework transferred clinically useful signal from an adult development cohort to a real-world pediatric dataset with monogenic movement disoders, without backbone retraining. Lightweight local calibration of the final decision layer substantially improved performance in the external pediatric cohort. The central contribution of this work is methodological and strategic: by separating a reusable shared backbone from a locally adapted decision layer, the framework avoids the full retraining burden that currently limits the clinical deployment of video-AI systems in rare and heterogeneous MDs. The calibration requirement, as few as five labeled patients per site, is compatible with the annotated caselads available in specialist centers, making onboarding realistic in practice. Calibration most clearly recovered chorea and athetosis, which were undetected at baseline, while dystonia, the dominant phenomenology, was reliably detected throughout; gains were more limited for low-prevalence or low-agreement labels such as myoclonus. This phenotype specificity is clinically informative: it identifies where the current framework is deployment-ready and where further development, richer visual encoding, 3D tracking, larger training cohorts, is needed before clinical integration. Validated in larger prospective multi-center cohorts, this two-stage architecture, reusable backbone plus site-specific calibration, could help move digital phenotyping of MDs from narrowly controlled proof-of-concept studies toward scalable, lifespan-spanning clinical deployment, supporting specialist practice without requiring specialist expertise at every point of care and importantly in AI.

\end{document}